\documentclass[runningheads]{llncs}
\usepackage{graphicx}
\usepackage{svg}
\usepackage{amsmath}
\usepackage{booktabs}
\usepackage{multirow}
\usepackage{array}
\usepackage{hyperref}
\usepackage{rotating}
\usepackage{tabularx}
\usepackage{url}
\newcolumntype{P}[1]{>{\centering\arraybackslash}p{#1}}
\newcolumntype{Y}{>{\centering\arraybackslash}X}
\begin{document}

\title{X-Ray to CT Rigid Registration Using \\ Scene Coordinate Regression}

\author{
%index{Shrestha, Pragyan}
Pragyan Shrestha\inst{1} \and
%index{Xie, Chun}
Chun Xie\inst{1} \and
%index{Shishido, Hidehiko}
Hidehiko Shishido\inst{1} \and \\
%index{Yoshii, Yuichi}
Yuichi Yoshii\inst{2} \and
%index{Kitahara, Itaru}
Itaru Kitahara\inst{1} 
}
\authorrunning{P. Shrestha et al.}
\institute{University of Tsukuba, Tsukuba, Ibaraki, Japan 
\email{shrestha.pragyan@image.iit.tsukuba.ac.jp}
\email{\{xiechun,shishido,kitahara\}@ccs.tsukuba.ac.jp}
\and
Tokyo Medical University, Ami, Ibaraki, Japan\\
\email{yyoshii@tokyo-med.ac.jp}
}
\maketitle
\begin{abstract}
Intraoperative fluoroscopy serves as a frequently employed modality in minimally invasive orthopedic surgeries. Aligning the intraoperatively acquired X-Ray image with the preoperatively acquired 3D model of a computed tomography (CT) scan reduces the mental burden on surgeons induced by the overlapping anatomical structures in the acquired images.  This paper proposes a fully automatic registration method that is robust to extreme viewpoints and does not require manual annotation of landmark points during training. It is based on a fully convolutional neural network (CNN) that regresses scene coordinates for a given X-Ray image. Scene coordinates are defined as the intersection of the back-projected ray from a pixel toward the 3D model. Training data for a patient-specific model is generated through a realistic simulation of a C-arm device using preoperative CT scans while intraoperative registration is achieved by solving the perspective-n-point (PnP) problem with random sample and consensus (RANSAC) algorithm. Experiments were conducted using a pelvis CT dataset including several real fluoroscopic (X-Ray) images with ground truth annotations. The proposed method achieved an average mean target registration error (mTRE) of 3.79+/1.67 mm in the 50\textsuperscript{th} percentile of the simulated test dataset and projected mTRE of 9.65+/-4.07 mm in the 50\textsuperscript{th} percentile of real fluoroscopic images for pelvis registration. Code is available at \url{https://github.com/Pragyanstha/SCR-Registration}.

\end{abstract}
\section{Introduction}
Image-guided navigation plays a crucial role in modern surgical procedures. In the field of orthopedics, many surgical procedures such as total hip arthroplasty, total knee arthroplasty, and pedicle screw injections utilize intraoperative fluoroscopy for surgical navigation \cite{Belei2007-nt,Bradley2019-qv,Merloz2007-xs}. Due to overlapping anatomical structures in X-Ray images, it is often difficult to correctly identify and reason the 3D structure from solely the image. Therefore, registration of an intraoperatively acquired X-Ray image to the preoperatively acquired CT scan is crucial in performing such procedures \cite{Woerner2016-qx,Selles2020-vo,Wylie2017-vr,Reichert2022-gg}. The standard procedure for acquiring highly accurate registration involves embedding fiducial markers into the patient and acquiring a preoperative CT scan \cite{Tang2000-yq,Maurer1997-rl,George2011-ep}. Intraoperative registration is performed using the explicitly identified 2D-3D correspondences. Inserting fiducial markers onto the body involves extra surgical costs and might not be a viable option for minimally invasive surgeries. To circumvent such issues with the feature-based method, an intensity-based optimization scheme for registration has been extensively researched in the past \cite{Aouadi2008-ja,Livyatan2003-hc}. Since the objective function is highly nonlinear for optimizing pose parameters, a good initialization is necessary for the method to converge in a global minimum. Therefore, it is usually accompanied by initial coarse registration using manual alignment of the 3D model to the image, which interrupts the surgical flow. On the other hand, learning-based methods have proved to be efficient in solving the registration task. Existing learning-based methods can be categorized broadly into two subdomains: landmark estimation and direct pose regression. Landmark estimation methods aim to solve for pose using correspondences between 3D landmark annotations and its estimated 2D projection points \cite{Grupp2020-xz,Bier2018-ht,Esteban2019-zg}, while methods based on pose regression estimate the global camera pose in a single inference \cite{Miao2015-wa}. Pose regressors are known to overfit training data and generalize poorly to unseen images \cite{Sattler2019-lw}. This makes the landmark estimation methods stand out in terms of registration quality as well as generalization. However, there exist two main issues with landmark estimation methods: 1) Annotation cost of a sufficiently large number of landmarks in the CT image. 2) Failure to solve for the pose in extreme views where projected landmarks are not visible or the number of visible landmarks is small. \\
In this paper, these issues are addressed by introducing scene coordinates \cite{Shotton2013-fn} to establish dense 2D-3D correspondences. Specifically, the proposed method regresses the scene coordinates of the CT-scan model from corresponding X-Ray images. A rigid transformation that aligns the CT-scan model to the image is then calculated by solving the PnP problem with the RANSAC algorithm.
\begin{figure}[t]
    \centering
    \includegraphics[width=0.8\textwidth]{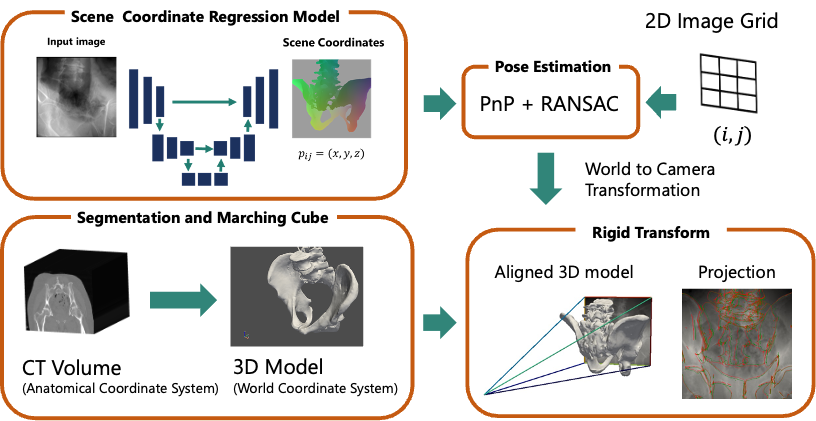}
    \caption{An overview of the proposed method. Scene coordinates are regressed using a U-Net architecture given an X-Ray image. With the obtained dense correspondences, PnP with RANSAC is run to get the transformation matrix that aligns the projection of the 3D model with the X-Ray image in the camera coordinate system.}
    \label{method}
\end{figure}
\section{Method}
\subsection{Problem Formulation}
The problem of 2D-3D registration can be formulated as finding the rigid transformation that transforms the 3D model defined in the anatomical or world coordinate system to the camera coordinate system. Specifically, given a CT-scan volume \(V_{CT} (x_w)\) where \(x_w\) is defined in the world coordinate system, the registration problem is concerned with finding \(T_w^c=[R|t]\) such that the following holds. 

\begin{equation}
\label{eq:1}
    I=\Re\{V_{CT} (T_w^{c^{-1}} x_c );K\}
\end{equation}

Where $\Re\{\cdot\}$ is the X-Ray transform that can be applied to volumes in the camera coordinate system given an intrinsic matrix $K$ and $I$, the target X-Ray image.

\subsection{Registration}
The overview of the proposed registration pipeline is shown in Fig. \ref{method}. The proposed method has four parts. First, the scene coordinates are regressed given a single-view X-Ray image as the input to a U-Net model. Second, the PnP + RANSAC algorithm is used to solve for the pose of the captured X-Ray system. Third, the CT-scan volume is segmented to obtain a 3D model of the bone regions. And fourth, the computed rigid transformation from world coordinates to camera coordinates is used to generate projection overlay images.

\subsubsection{Scene Coordinates.}
Scene coordinates are defined as the points of intersection between a camera’s back-projected rays and the 3D model in a world coordinate system (i.e., only the first intersection and the last intersection are considered). The same concept is adapted for X-Ray images and its underlying 3D model obtained from CT-scan. Specifically, given an arbitrary point $x_{ij}$ in the image plane, the scene coordinates $X_{ij}$ satisfy the following conditions.
\begin{equation}
\label{eq:2}
    X_{ij}=[R^T |-t](d K^T x_{ij})
\end{equation}
where $R$ and $t$ are the rotation matrix and translation vector that maps points in the world coordinate system to the camera coordinate system, $K$ is the intrinsic matrix, $d$ is the depth, as seen from the camera, of the point $X$ on the 3D model.

\subsubsection{Uncertainty Estimation.}
The task of scene coordinate regression is to estimate these $X_{ij}$ for every pixel $ij$, given an X-Ray image $I$. However, the existence of $X_{ij}$ is not guaranteed for all pixels since the back-projected rays may not intersect the 3D model. One of many ways to address such a case would be to prepare a mask (i.e., 1 if bone area, 0 otherwise) in advance so that only the pixels that lie inside the mask are estimated. Since this approach would require an explicit way of estimating the mask image, an alternative approach has been taken in this work. Instead of estimating a single $X_{ij}$, the mean and variance of the scene coordinate is estimated. The non-intersecting scene coordinates are identified by applying thresholding to the estimated variance (i.e., points with high variance are considered non-existent scene coordinates and thus filtered out). This approach assumes that the observed scene coordinates are corrupted with a zero mean, non-zero and non-constant variance, and isotropic Gaussian noise.
\begin{equation}
\label{eq:3}
    X_{ij}\sim N(u(I,x_{ij} ),\sigma(I,x_{ij}))
\end{equation}
Where $u(I,x_{ij})$ and $\sigma(I,x_{ij})$ are the functions producing the mean and standard deviation of the scene coordinate respectively. This work represents these functions using a fully convolutional neural network.
\subsubsection{Loss Function.}
A U-Net architecture is used for estimating the mean and standard deviation of scene coordinates at every pixel in a given image. The loss function for intersecting scene coordinates is derived from the maximum likelihood estimates using the likelihood $X_{ij}$. It can be written as follows:
\begin{equation}
\label{eq:4}
    Loss_{intersecting}=(\frac{(X_{ij}-u(I,x_{ij}))}{\sigma(I,x_{ij} )})^2+2\log(\sigma(I,x_{ij}))
\end{equation}

Since it is desired to have a high variance for non-existent scene coordinates, the loss function for non-existent coordinates is designed as follows:
\begin{equation}
\label{eq:5}
    Loss_{non-existent}=\frac{1}{\sigma(I,x_{ij})}
\end{equation}
\subsubsection{2D-3D Registration.}
Iterative PnP implementation from OpenCV is run using RANSAC with maximum iteration of 1000 and reprojection error of 10px and 20px for simulated and real X-Ray images respectively. An example of a successful registration is shown in the left part of Fig. \ref{example_successfull_dataset} below.
\begin{figure}
    \centering
\includegraphics[width=0.9\columnwidth]{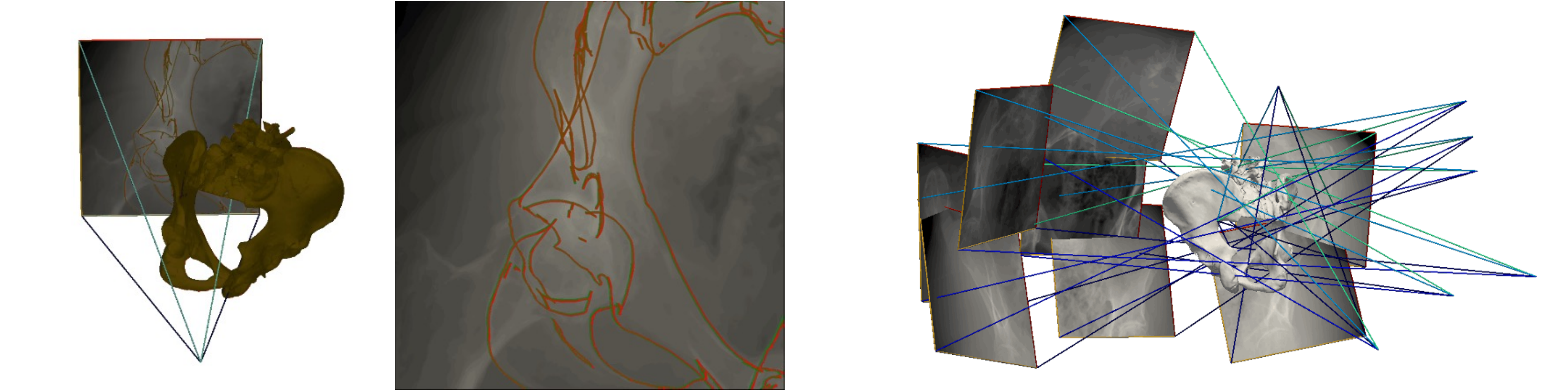}
    \caption{An example of successful registration with the proposed method (left two images) and Randomly picked data samples in the test set (right). The X-Ray image and model’s gradient projection overlay (middle) and the pose of the model in the camera coordinates system (left). The origin of the view frustum is the X-Ray source position and the simulated X-Ray images are placed in the detector plane for visualization (right).}
    \label{example_successfull_dataset}
\end{figure}
\section{Experiments and Results}
\subsection{Dataset}
To properly evaluate the proposed method, a dataset that contains 6 annotated CT scans each with several registered real X-Ray images from \cite{Grupp2020-xz} was used. The annotations include 14 landmarks and 7 segmentation labels. The CT scans are pelvic bones of cadaveric specimens. Since there are only a few real X-Ray images, simulated X-Ray images are generated from each CT-scans for training and testing the model. Specifically, DeepDRR \cite{Bier2018-ht} was used for simulating a Siemens Cios Fusion Mobile C-arm imaging device. Similar to \cite{Bier2018-ht}, LAO/RAO of [-45, 45] degrees respectively were samples at 1-degree intervals. Random offset was added in each direction. The offset vector was sampled from a normal distribution of zero means and 90 mm standard deviation in the lateral direction, and 30 mm standard deviations each in the other two directions. Through this, images with partially visible structures were intentionally simulated. Some randomly picked samples are shown in the right part of Fig. \ref{example_successfull_dataset}. For each image, the ground truth scene coordinates were obtained from the 3D model of the CT scans. For each specimen, 8100 simulated X-Rays were generated, of which, 5184 images were randomly assigned as the training set, 1296 for the validation set, and the remaining 1620 for the test set. 

\subsection{Implementation Details}
The U-Net was implemented in Pytorch 1.13.0. An image size of 512×512 is used for input as well as output scene coordinates. The output channel size is 8 (i.e., 3 for scene coordinates and 1 for standard deviation, multiplied by 2 for entry and exit points). The model was trained with each dataset individually (i.e., patient-specific models were obtained for each specimen) using Adam with a constant learning rate of 0.0001 and batch size of 16. Online data augmentation with a probability of 0.5 was applied for domain randomization. It included random invert, color jitter with brightness and contrast parameters each set to 1, and random erasing. The scene coordinates were filtered using a log variance threshold of 0 for simulated images and -2 for real X-Ray images.

\subsection{Baselines and Evaluation Metrics}
The proposed method was compared against two other baseline methods: PoseNet \cite{Kendall2015-mk} and DFLNet \cite{Grupp2020-xz}. PoseNet was implemented using ResNet-50 as the backbone for the feature extractor and trained using geometric loss. DFLNet uses the same architecture as the proposed method however the last layer regresses 14 heatmaps of the landmarks instead of scene coordinates. Note that the segmentation layer along with the gradient-based optimization phase in the original paper has been left out for architectural comparison. Each baseline was trained in a patient-specific manner following the proposed method. The mean target registration error (mTRE) and Gross Failure Rate (GFR), were used as the evaluation metric to compare with the baselines. mTRE is defined in \ref{eq:6}, where \(X_k\) is the position of the ground truth landmark \(\hat{X_k}\) after applying the predicted transformation. GFR is the ratio of failed cases where the failed cases are defined as the registration results with mTRE greater than 10mm. Since we could only get the projection of ground truth landmarks and not the ground truth transformation matrix for the real X-Ray images, projected mTRE (proj. mTRE) was used for evaluation. It is similar to mTRE except the \(X_k\) and \(\hat{X_k}\) represent the projected coordinates of the landmarks, in the detector plane (i.e., the pixel coordinates are scaled according to the detector size to match the units).
\begin{equation}
\label{eq:6}
    \mathrm{mTRE} =\frac{1}{N}\sum_{k=1}^{k=N}\|X_k - \hat{X_k}\|_2
\end{equation}

\subsection{Registration Results}
\subsubsection{Simulated X-Ray Images.}
\begin{figure}[t]
    \centering
\includegraphics[width=1.0\columnwidth]{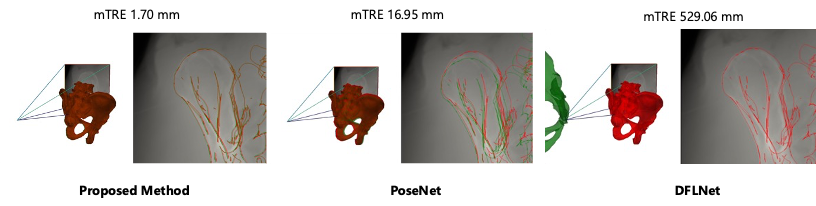}
    \caption{An example case illustrating an extreme partial viewpoint. The proposed method successfully registers the image with 1.70 mm mTRE, while PoseNet struggles with 16.95 mm mTRE. Since there is an insufficient number (less than 4) of visible landmarks, the DFLNet hallucinates landmarks providing incorrect 2D-3D correspondences which leads to large mTRE.}
    \label{extreme_views}
\end{figure}

Table \ref{eval_sim} shows the mTRE in the 25\textsuperscript{th}, 50\textsuperscript{th}, and 95\textsuperscript{th} percentile of the total test sample size and the GFR. For most of the specimens, the proposed method could retain the GFR below 20\% whereas PoseNet and DFLNet fail to register with more than 20\% GFR in most cases. For PoseNet, this is because the network cannot reason about the spatial structure and its local relation to the image patches. For DFLNet, this is inevitable due to the visibility issue of landmark points that were mostly located in the pubic region of the pelvis. Comparing the mTRE of each specimen with each method, the proposed method achieved an mTRE of 7.98 mm even in the 95\textsuperscript{th} percentile of specimen 2. DFLNet achieved the lowest mTRE of 0.98 mm in the 25\textsuperscript{th} percentile of specimen 4. This illustrates the highly accurate registration that landmark estimation methods are capable of. However, with extreme or partial views such as the one shown in Fig. \ref{extreme_views}, the method cannot estimate the correct pose parameter due to incorrect landmark localization or an insufficient number of visible landmarks. Please refer to the supplemental material for registration overlay results of different specimens using the proposed method. 

\begin{table}
\tiny
    \centering
    \caption{The mean target registration errors each in $25^{th}$, $50^{th}$, and $95^{th}$ percentile of the simulated test dataset. All models are trained individually on the 6 specimens shown below. The proposed method outperforms other methods in terms of $50^{th}$ percentile mTRE and GFR in most of the specimens.}
    \label{eval_sim}
    \begin{tabularx}{\textwidth}{c *{13}{Y}}
    \toprule
     {}
     & \multicolumn{4}{c}{PoseNet}  
     & \multicolumn{4}{c}{DFLNet}
     & \multicolumn{4}{c}{Ours}\\
        \cmidrule(lr){2-5} \cmidrule(lr){6-9} \cmidrule(lr){10-13}
     {} & \multicolumn{3}{c}{mTRE[mm]$\downarrow$} & {} & \multicolumn{3}{c}{mTRE[mm]$\downarrow$} & {} & \multicolumn{3}{c}{mTRE[mm]$\downarrow$} & {} \\
     Specimen & $25^{th}$ & $50^{th}$ & $95^{th}$ & GFR[$\%$]$\downarrow$ & $25^{th}$ & $50^{th}$ & $95^{th}$ & GFR[$\%$]$\downarrow$ & $25^{th}$ & $50^{th}$ & $95^{th}$ & GFR[$\%$]$\downarrow$ \\
    \midrule

    \#1 & 5.75 & 8.37 & 24.81 & 38.53 & 3.36 & 212.59 & 680.58 & 62.27 & 1.37 & \textbf{2.50} & 9.80 & \textbf{4.87} \\
    \#2 & 6.97 & 10.23 & 25.95 & 51.63 & 1.98 & 7.04 & 656.41 & 46.15 & 1.15 & \textbf{2.14} & 7.98 & \textbf{2.54} \\
    \#3 & 5.42 & 7.86 & 23.34 & 35.35 & 1.03 & \textbf{2.51} & 583.63 & 28.65 & 1.67 & 3.05 & 12.25 & \textbf{8.56} \\
    \#4 & 4.67 & 6.46 & 16.91 & 18.77 & 0.98 & \textbf{2.30} & 558.20 & 23.59 & 1.76 & 3.38 & 19.19 & \textbf{12.54} \\
    \#5 & 4.81 & 6.52 & 18.98 & 22.43 & 1.51 & \textbf{4.28} & 767.56 & 37.47 & 3.09 & 5.30 & 17.32 & \textbf{18.85} \\
    \#6 & 4.06 & \textbf{5.85} & 18.42 & \textbf{22.69} & 2.26 & 139.96 & 15321.19 & 58.72 & 3.80 & 6.37 & 18.25 & 23.18 \\        
    \hline
    \addlinespace[3pt]
    mean & 5.28 & 7.55 & 21.40 & 31.57 & 1.85 & 61.45 & 3094.60 & 42.81 & 2.14 & \textbf{3.79} & 14.13 & \textbf{11.76} \\
    std & 1.02 & 1.62 & 3.77 & 12.58 & 0.90 & 91.88 & 5990.24 & 15.76 & 1.06 & 1.67 & 4.75 & 8.05 \\
    \bottomrule
    \end{tabularx}
\end{table}

\subsubsection{Real X-Ray Images.}
Table \ref{eval_real} shows the mTRE calculated on projected image points (abbreviated as proj. mTRE) for PoseNet and the proposed method. DFLNet could not adapt to real X-Ray images, therefore, was left out of the table. Since our dataset consisted mostly of images with partially visible hips, only a few landmarks are visible per image. This causes the DFLNet to overfit to the partially visible landmark distribution while our proposed model mitigates this issue by learning the general structure (i.e., every surface point that is visible). The proposed method estimated good transformations (i.e., proj. mTRE approximately around 10 mm in the 50\textsuperscript{th} percentile). In contrast, the proj. mTRE for PoseNet is significantly higher. This suggests that PoseNet overfits the training data despite the application of domain randomization. This result agrees with previous reports \cite{Sattler2019-lw} that address this issue. A visualization of the overlays is presented in the supplemental material.

\begin{table}
\tiny
    \centering
    \caption{The projective mean target registration error evaluated on real X-Ray dataset. The proposed method achieved significantly low registration errors compared to PoseNet, implying that it generalizes well to unseen data and domain.}
  \label{eval_real}
    \begin{tabular}{ P{1.5cm} P{1.5cm} P{1.4cm} P{1.4cm} P{1.4cm} P{1.4cm} P{1.4cm} P{1.4cm} }
     \toprule
     {} & {} &
     \multicolumn{3}{c}{PoseNet} &
     \multicolumn{3}{c}{Ours} \\
     \cmidrule(lr){3-5} \cmidrule(lr){6-8}
     {} & Number \newline of & \multicolumn{3}{c}{proj. mTRE [mm]$\downarrow$} & \multicolumn{3}{c}{proj. mTRE [mm]$\downarrow$} \\
     Specimen & Images & $25^{th}$ & $50^{th}$ & $95^{th}$ & $25^{th}$ & $50^{th}$ & $95^{th}$ \\
     \midrule
     \#1 & 111 & 43.64 & 49.11 & 64.23 & 5.45 & \textbf{8.02} & 55.87 \\
     \#2 & 24 & 19.42 & 27.18 & 43.68 & 2.74 & \textbf{3.32} & 6.48 \\
     \#3 & 104 & 31.18 & 38.97 & 66.06 & 7.60 & \textbf{11.85} & 162.83 \\
     \#4 & 24 & 35.52 & 38.37 & 57.07 & 11.12 & \textbf{15.52} & 92.34 \\
     \#5 & 48 & 38.97 & 46.60 & 69.06 & 6.09 & \textbf{9.07} & 21.91 \\
     \#6 & 55 & 34.51 & 37.13 & 47.72 & 7.18 & \textbf{10.14} & 20.78 \\
    \hline
    \addlinespace[3pt]
     mean & {} & 33.87 & 39.56 & 57.97 & 6.70 & \textbf{9.65} & 60.04  \\
     std & {} & 8.25 & 7.77 & 10.37 & 2.77 & 4.07 & 59.15 \\
     \bottomrule
    \end{tabular}
\end{table}

\section{Limitations}
As the proposed method is designed for producing initial estimates of the pose parameters, an extra refinement step using an intensity-based optimization method is required for obtaining clinically relevant registration accuracy. Although the proposed method can provide a good initial estimate, the average runtime for the whole pipeline was 1.75 seconds which is around two orders of magnitude greater than PoseNet, which had an average runtime of 0.06 seconds. This is because RANSAC has to find a good pose from a dense set of correspondences. This issue may be addressed by heuristically selecting a good variance threshold per image that filters out bad correspondences. 

\section{Conclusion}
This paper presented a scene coordinate regression-based approach for the X-Ray to CT-scan model registration problem. The proposed method does not require labeling of anatomical landmarks and is effective in extreme view angles. Experiments with simulated X-Ray images, as well as real X-Ray images, showed that the proposed method could perform well even under partially visible structures and extreme view angles, compared to direct pose regression and landmark estimation methods. Testing the model trained solely on simulated X-Ray images, on real X-Ray images did not result in catastrophic failure, instead, the results were positive for instantiating further refinement steps.

\subsubsection{Acknowledgement}
This work was partially supported by a grant from JSPS KAKENHI Grant Number JP23K08618. This work (in part) used computational resources of Cygnus provided by Multidisciplinary Cooperative Research Program in Center for Computational Sciences, University of Tsukuba.

\bibliographystyle{splncs04}
\bibliography{references}

\begin{thebibliography}{10}
\providecommand{\url}[1]{\texttt{#1}}
\providecommand{\urlprefix}{URL }
\providecommand{\doi}[1]{https://doi.org/#1}

\bibitem{Aouadi2008-ja}
Aouadi, S., Sarry, L.: Accurate and precise {2D--3D} registration based on
  x-ray intensity. Comput. Vis. Image Underst.  \textbf{110}(1),  134--151 (Apr
  2008)

\bibitem{Belei2007-nt}
Belei, P., Skwara, A., De~La~Fuente, M., Schkommodau, E., Fuchs, S., Wirtz,
  D.C., K{\"a}mper, C., Radermacher, K.: Fluoroscopic navigation system for hip
  surface replacement. Comput. Aided Surg.  \textbf{12}(3),  160--167 (May
  2007)

\bibitem{Bier2018-ht}
Bier, B., Unberath, M., Zaech, J.N., Fotouhi, J., Armand, M., Osgood, G.,
  Navab, N., Maier, A.: X-ray-transform invariant anatomical landmark detection
  for pelvic trauma surgery  (Mar 2018)

\bibitem{Bradley2019-qv}
Bradley, M.P., Benson, J.R., Muir, J.M.: Accuracy of acetabular component
  positioning using computer-assisted navigation in direct anterior total hip
  arthroplasty. Cureus  \textbf{11}(4),  e4478 (Apr 2019)

\bibitem{Esteban2019-zg}
Esteban, J., Grimm, M., Unberath, M., Zahnd, G., Navab, N.: Towards fully
  automatic {X-Ray} to {CT} registration. In: Medical Image Computing and
  Computer Assisted Intervention -- {MICCAI} 2019. pp. 631--639. Springer
  International Publishing (2019)

\bibitem{George2011-ep}
George, A.K., Sonmez, M., Lederman, R.J., Faranesh, A.Z.: Robust automatic
  rigid registration of {MRI} and x-ray using external fiducial markers for
  {XFM-guided} interventional procedures. Med. Phys.  \textbf{38}(1),  125--141
  (Jan 2011)

\bibitem{Grupp2020-xz}
Grupp, R.B., Unberath, M., Gao, C., Hegeman, R.A., Murphy, R.J., Alexander,
  C.P., Otake, Y., McArthur, B.A., Armand, M., Taylor, R.H.: Automatic
  annotation of hip anatomy in fluoroscopy for robust and efficient {2D/3D}
  registration. Int. J. Comput. Assist. Radiol. Surg.  \textbf{15}(5),
  759--769 (May 2020)

\bibitem{Kendall2015-mk}
Kendall, A., Grimes, M., Cipolla, R.: {PoseNet}: A convolutional network for
  {Real-Time} {6-DOF} camera relocalization. In: 2015 {IEEE} International
  Conference on Computer Vision ({ICCV}). pp. 2938--2946 (Dec 2015)

\bibitem{Livyatan2003-hc}
Livyatan, H., Yaniv, Z., Joskowicz, L.: Gradient-based {2-D/3-D} rigid
  registration of fluoroscopic x-ray to {CT}. IEEE Trans. Med. Imaging
  \textbf{22}(11),  1395--1406 (Nov 2003)

\bibitem{Maurer1997-rl}
Maurer, Jr, C.R., Fitzpatrick, J.M., Wang, M.Y., Galloway, Jr, R.L., Maciunas,
  R.J., Allen, G.S.: Registration of head volume images using implantable
  fiducial markers. IEEE Trans. Med. Imaging  \textbf{16}(4),  447--462 (Aug
  1997)

\bibitem{Merloz2007-xs}
Merloz, P., Troccaz, J., Vouaillat, H., Vasile, C., Tonetti, J., Eid, A.,
  Plaweski, S.: Fluoroscopy-based navigation system in spine surgery. Proc.
  Inst. Mech. Eng. H  \textbf{221}(7),  813--820 (Oct 2007)

\bibitem{Miao2015-wa}
Miao, S., Jane~Wang, Z., Liao, R.: Real-time {2D/3D} registration via {CNN}
  regression  (Jul 2015)

\bibitem{Reichert2022-gg}
Reichert, J.C., Hofer, A., Matziolis, G., Wassilew, G.I.: Intraoperative
  fluoroscopy allows the reliable assessment of deformity correction during
  periacetabular osteotomy. J. Clin. Med. Res.  \textbf{11}(16) (Aug 2022)

\bibitem{Sattler2019-lw}
Sattler, T., Zhou, Q., Pollefeys, M., Leal-Taix{\'e}, L.: Understanding the
  limitations of {CNN-Based} absolute camera pose regression. In: 2019
  {IEEE/CVF} Conference on Computer Vision and Pattern Recognition ({CVPR}).
  pp. 3297--3307. unknown (Jun 2019)

\bibitem{Selles2020-vo}
Selles, C.A., Beerekamp, M.S.H., Leenhouts, P.A., Segers, M.J.M., Goslings,
  J.C., Schep, N.W.L., {EF3X Study Group}: The value of intraoperative
  3-dimensional fluoroscopy in the treatment of distal radius fractures: A
  randomized clinical trial. J. Hand Surg. Am.  \textbf{45}(3),  189--195 (Mar
  2020)

\bibitem{Shotton2013-fn}
Shotton, J., Glocker, B., Zach, C., Izadi, S., Criminisi, A., Fitzgibbon, A.:
  Scene coordinate regression forests for camera relocalization in {RGB-D}
  images. In: 2013 {IEEE} Conference on Computer Vision and Pattern
  Recognition. pp. 2930--2937 (Jun 2013)

\bibitem{Tang2000-yq}
Tang, T.S.Y., Ellis, R.E., Fichtinger, G.: Fiducial registration from a single
  {X-Ray} image: A new technique for fluoroscopic guidance and radiotherapy.
  In: Medical Image Computing and {Computer-Assisted} Intervention -- {MICCAI}
  2000. pp. 502--511. Springer Berlin Heidelberg (2000)

\bibitem{Woerner2016-qx}
Woerner, M., Sendtner, E., Springorum, R., Craiovan, B., Worlicek, M.,
  Renkawitz, T., Grifka, J., Weber, M.: Visual intraoperative estimation of cup
  and stem position is not reliable in minimally invasive hip arthroplasty.
  Acta Orthop.  \textbf{87}(3),  225--230 (Jun 2016)

\bibitem{Wylie2017-vr}
Wylie, J.D., Ross, J.A., Erickson, J.A., Anderson, M.B., Peters, C.L.:
  Operative fluoroscopic correction is reliable and correlates with
  postoperative radiographic correction in periacetabular osteotomy. Clin.
  Orthop. Relat. Res.  \textbf{475}(4),  1100--1106 (Apr 2017)

\end{thebibliography}

\end{document}